\documentclass[journal]{IEEEtran}
\usepackage{graphicx}
\usepackage{cite}
\usepackage{picinpar}
\usepackage{amsmath}
\usepackage{url}
\usepackage{flushend}
\usepackage{algorithm}
\usepackage{algpseudocode}
\usepackage[latin1]{inputenc}
\usepackage{colortbl}
\usepackage{soul}
\usepackage{multirow}
\usepackage{pifont}
\usepackage{color}
\usepackage{alltt}
\usepackage[hidelinks]{hyperref}
\usepackage{enumerate}
\usepackage{siunitx}
\usepackage{breakurl}
\usepackage{epstopdf}
\usepackage{pbox}

\begin{document}
\title{Meta Learning for Task-Driven Video Summarization}

\author{
	\vskip 1em
	{
	Xuelong Li, \emph{Fellow, IEEE},
	Hongli Li, and Yongsheng Dong, \emph{Senior Member, IEEE}
	}

	\thanks{
		
		{
	
This work was supported in part by the National Key Research and Development Program of China under Grant 2018YFB1107400, in part by the National Natural Science Foundation of China under Grant 61871470 and Grant U1604153, in part by the Key Specialized Research and Development Breakthrough of Henan Province under Grant 192102210121, and in part by the Program for	Science and Technology Innovation Talents in Universities of Henan Province under Grant 19HASTIT026.
		
X. Li and H. Li are with the School of Computer Science and the Center for OPTical IMagery Analysis and Learning (OPTIMAL), Northwestern Polytechnical University, Xi'an 710072, P.R. China (e-mails: xuelong\_li@nwpu.edu.cn; honglili0428@gmail.com).
			
Y. Dong is with the School of Information Engineering, Henan University of Science and Technology, Luoyang 471023, China (e-mail: dongyongsheng98@163.com).
		}
	}
}

\maketitle
	
\begin{abstract}
Existing video summarization approaches mainly concentrate on sequential or structural characteristic of video data. However, they do not pay enough attention to the video summarization task itself.
In this paper, we propose a meta learning method for performing task-driven video summarization, denoted by MetaL-TDVS, to explicitly explore the video summarization	mechanism among summarizing processes on different videos. Particularly, MetaL-TDVS aims to excavate the latent mechanism for summarizing video by reformulating video summarization as a meta learning problem and promote generalization ability of the trained model. MetaL-TDVS	regards summarizing each video as a single task to make better use of the experience and knowledge learned from processes of summarizing other videos to summarize new ones.
Furthermore, MetaL-TDVS updates models via a two-fold back propagation which forces the model optimized on one video to obtain high accuracy on another video in every training step.
Extensive experiments on benchmark datasets demonstrate the superiority and better generalization ability of MetaL-TDVS against several state-of-the-art methods.
\end{abstract}


\markboth{IEEE TRANSACTIONS ON INDUSTRIAL ELECTRONICS}%
{}

\definecolor{limegreen}{rgb}{0.2, 0.8, 0.2}
\definecolor{forestgreen}{rgb}{0.13, 0.55, 0.13}
\definecolor{greenhtml}{rgb}{0.0, 0.5, 0.0}
\section{Introduction}
\label{intro}
\IEEEPARstart{H}{umans} get information from videos easily, but the time cost to browse these videos is noticeable. Thus, an efficient way to manage large amount of videos is needed urgently\cite{zhang2015effective,ji2018hypergraph,ji2019query}. Being able to capture the essence and discard redundant frames in videos automatically, video summarization meets this requirement well \cite{ji2019multi}, and is widely used in video storage, displaying, compression etc. For instance, due to limited battery energy and unstable internet speed, it is better to transmit a representative summary instead of entire video for some real-time communication mobile phone apps.

Unsupervised video summarization methods \cite{panda2017collaborative,zhao2014quasi} usually use manually defined criteria to extract key frames or key shots. While supervised ones \cite{plummer2017enhancing,zhao2018hsa} learn models with the help of human-annotated data to determine which frames or shots are more important. In this paper, we mainly focus on supervised ones.

The majority of existing supervised methods mainly pay more attention to sequential or structural nature of video data. Zhang \textit{et al.} stated that video summarization is inherently a structured prediction problem and proposed a supervised subset selection technique \cite{zhang2016summary} to transfer summary structures from training labeled videos to unseen ones. Though this approach gets attractive results, its assumption, that is similar videos have similar summary structures, requires scenes of training video to be abundant enough. Obviously, when training data is not sufficiently rich, generalization ability of the learned model will be limited.

To model temporal dependency among video frames in variable range, two long short term memory (LSTM) based models were proposed by casting video summarization task as a structured prediction problem \cite{zhang2016video}. For computing probability of each frame being selected into summary, a deep summarization network (DSN) was constructed by viewing summarizing video as a sequential decision making problem \cite{zhou2017deep}. To catch long range temporal dependency of video frames well, Zhao \textit{et al.} proposed a hierarchical RNN (H-RNN) \cite{zhao2017hierarchical}. Moreover, LSTM was also utilized in other methods \cite{ji2017video,mahasseni2017unsupervised} to model sequential or structural characteristics of video.

Though these existing supervised methods get state-of-the-art performance, they mainly concentrate on sequential or structural nature of video data rather than directly on video summarization task itself. In this way, the learned models also focus more on sequential or structural characteristic of video data, but not explicitly explore how to summarize video. Undoubtedly, sequential structure of video is critical to video summarization, but the mechanism for summarizing video is more crucial and actually essential to video summarization task itself.

Therefore, this work emphasizes on exploration of the latent mechanism for video summarization, and lays more stress on video summarization problem itself rather than only on video data. Specifically, we reformulate summarizing video as a meta learning problem and propose a general framework MetaL-TDVS to explicitly explore the latent mechanism for video summarization.

In MetaL-TDVS, summarizing each video is seen as a single task and learning proceeds among tasks. Specifically, parameter of learner (specific model to summarize video) is learned. Each update of parameter consists of two stages with two training tasks and parameter of learner can be obtained via a two-fold back propagation. As demonstrated by extensive experiments, MetaL-TDVS obtains better generalization ability and is superior than many state-of-the-art supervised methods.

Contributions of MetaL-TDVS are summarized as follows:

\begin{enumerate}
	\item We first propose a MetaL-TDVS method to perform video summarization by employing meta learning. To the best of our knowledge, it is the first one to use meta learning in video summarization domain.
	\item We use MetaL-TDVS to explicitly explore the latent mechanism of summarizing video, and focus more on video summarization problem itself instead of only on sequential or structural video data.
	\item We use both deep features and traditional features respectively to demonstrate the effectiveness of MetaL-TDVS. Experimental results reveal that MetaL-TDVS outperforms current representative video summarization methods, whether deep features or traditional features are used.
\end{enumerate}

Rest of the paper is organized as follows. Section \ref{related} talks about related works, section \ref{method} describes the proposed MetaL-TDVS in detail. Experimental details and results are shown and discussed in section \ref{experiment}, section \ref{conclusion} concludes the paper.

\section{Related Work}

\label{related}

Supervised video summarization methods rely on annotated data to train models and capture increasingly attention due to their outstanding performance. Among them, subset selection \cite{mahasseni2017unsupervised,gong2014diverse}, structured prediction \cite{zhang2016video}, sequential decision making \cite{zhou2017deep}, and sequence to sequence learning \cite{ji2017video} are four of the classical formulations.

A probabilistic model, sequential determinantal point process (seqDPP) \cite{gong2014diverse}, is designed to select a diverse and informative subset of items (video frames or shots) from ground set (whole video frames or shots). $\rm {SUM-GAN_{sup}}$ \cite{mahasseni2017unsupervised} includes a sLSTM (selector LSTM) which is devised and trained to be a subset selector and summary are generated from selected frames or shots. Two bidirectional LSTM based network models are proposed in \cite{zhang2016video} which can give either binary labels or importance scores of video frames when inputting frame features. $\rm {DR-DSN_{sup}}$ \cite{zhou2017deep} produces video summary via three steps: frame features are firstly generated by an encoder (a convolutional neural network); and decoder (a bidirectional LSTM network) computes probabilities according to the features; finally, approaches proposed in \cite{potapov2014category, song2015tvsum} are employed to make summary from probabilities. Encoder-decoder structure is also adopted in AVS \cite{ji2017video} where the encoder employs a bidirectional LSTM to extract features of video frames and the decoder uses two attention-based LSTMs to compute importance scores of frames.

Though existing supervised methods get promising results, nearly none of them explicitly explore the mechanism for summarizing video among summarizing processes on different videos. In contrast, meta learning is a choice worthy of consideration.

First proposed by Maudsley, meta learning is described as a process by which learners become aware of and increasingly taking control of their habits such as perception and learning \cite{maudsley1980theory}. After conceptual basis of meta learning set by Maudsely, Biggs interprets meta learning as a state of being aware of and in control of one's own learning \cite{biggs1985role}. Basically, meta learning can be interpreted as the process of learning to learn. Typically, in meta learning, meta-learner is a trainable algorithm \cite{li2017meta,andrychowicz2016learning} or a trainable model \cite{santoro2016meta,mishra2017meta} which can guide the learning of learner. Learner is a specific model which handles problems directly.

Imitating human ability of learning to learn at a high level, meta learning aims to make better use of the previously learned experience and knowledge to help learning on new tasks. Evidently, it is different with most supervised methods which learn each task in isolation and from scratch. By meta learning, the model trained among tasks can obtain ability of learning to learn and is able to learn new tasks quickly. Lifting learning level from data to task and being able to learn to learn, meta learning is a more intelligent way of learning, thus attracts growing attention. Following the idea of meta learning, there are many excellent studies. To be able to quickly adapt to a new task using only a few data-points and by a few training iterations, a model-agnostic meta-learning (MAML) algorithm was proposed in \cite{finn2017model}. A meta-learner called Meta-SGD was developed in \cite{li2017meta} for making better use of experience and knowledge learned from related tasks. Motivated by successful move from hand-designed features to learned features, a meta learning way to learn suitable optimization algorithms for some specific problems was proposed in \cite{andrychowicz2016learning}.

The idea of meta learning, that is not only learning but also learning to learn, seems tailored to video summarization problem. Because learning to summarize video (the mechanism for summarizing video) is more crucial than only learning to distinguish which frames are more important. Thus, video summarization problem itself is actually a problem of learning to learn (learn to summarize video) and meta learning is a reasonable way to address the video summarization problem.

\section{Proposed MetaL-TDVS}
\label{method}

This section firstly gives definition of MetaL-TDVS, then presents outline of MetaL-TDVS. Finally, a compact description of the specific video summarization model is introduced.

\subsection{Definition of MetaL-TDVS}
Video summarization is to summarize a given video by using the prior knowledge. The prior knowledge can be seen as meta knowledge, which can be obtained by meta learning from known videos. Suppose a video $v_i$ in video space $S_v$, summarizing $v_i$ is a single task $\tau_i$ in task space $S_\tau$ and summarizing different videos are seen as different tasks. As an example, summarizing $v_i$ is a single task $\tau_i$ while summarizing $v_j$ is another task $\tau_j$ if $v_i$ and $v_j$ are different. Based on this definition, summarizing all videos in any of video datasets (in this paper we use Youtube, OVP, TVSum, SumMe) form a task set $T\{\tau\}$. We follow rules in \cite{zhang2016video} to split $T\{\tau\}$ into three disjoint subsets: training task set $Tr\{\tau\}$ used for learning parameter of learner, validation task set $Val\{\tau\}$ that is used for deciding when learning can be stopped, testing task set $Te\{\tau\}$ used for computing performance of MetaL-TDVS.

Upon above settings, MetaL-TDVS can be defined as:
\begin{equation}
\setlength{\abovedisplayskip}{3pt}
\setlength{\belowdisplayskip}{3pt}
{learner^* = m\_learner(learner\_0, Tr\{\tau\}, Val\{\tau\}),}
\end{equation}
where $m\_learner$ denotes meta learner in meta learning. $learner\_0$ is the model of learner in meta learning and  is randomly initialized in our implementation. As specific model to summarize video, $learner\_0$ is implemented by vsLSTM \cite{zhang2016video}. In fact, it can be implemented by any differentiable model. $learner^*$ denotes optimized learner after learning.

\subsection{Details of MetaL-TDVS}
\label{MetaSum}

\begin{figure*}[t]
	\centering
	\includegraphics[width=18.5cm]{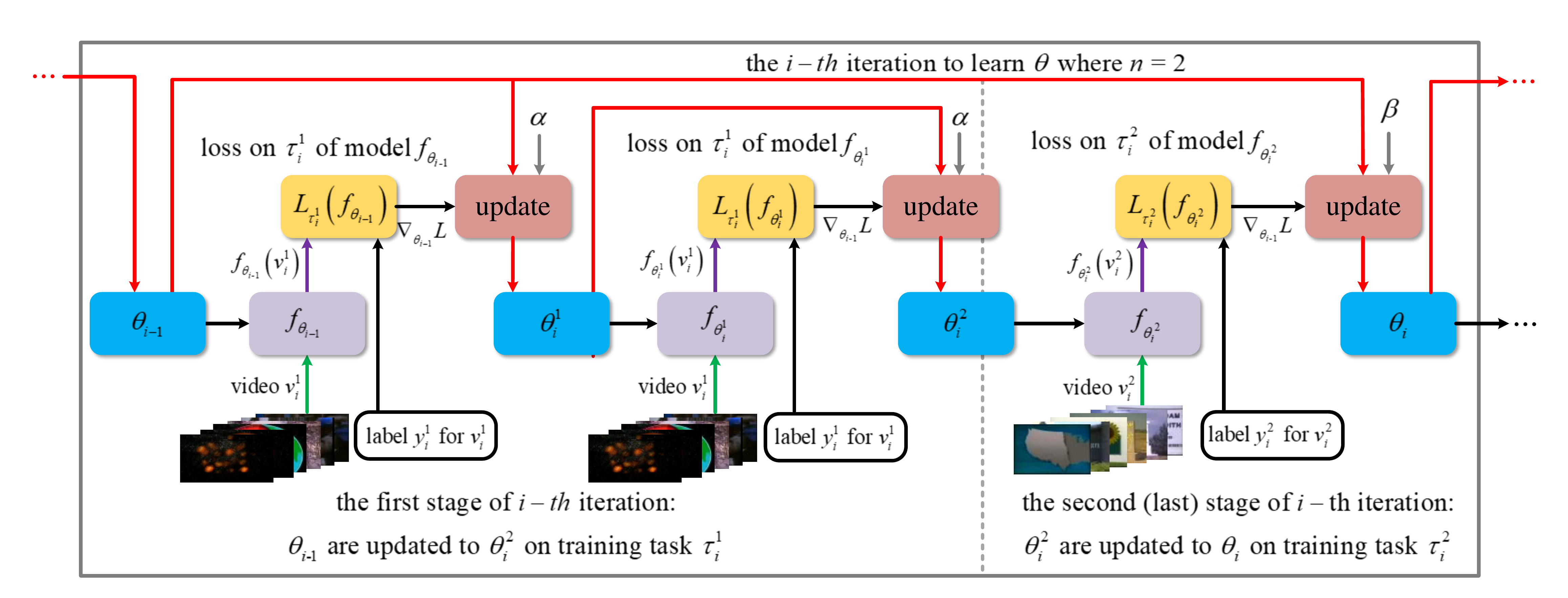}
	\setlength{\abovecaptionskip}{-0.65cm}
	\caption{Overview of the $i$-th iteration for update $\theta$ from $\theta_{i-1}$ to $\theta_{i}$ ($\theta$ is randomly initialized to $\theta_{0}$ at beginning): The update consists of two stages. The first stage updates $\theta_{i-1}$ to $\theta_{i}^{2}$ according to $\tau_i^1$ and the second stage updates $\theta_{i}^{2}$ to $\theta_{i}$ on $\tau_i^2$. Every ``update'' is done by one gradient descent step. In this example $n=2$. $\alpha$ denotes learning rate and $\beta$ is meta learning rate. Green arrows represent inputs of learner which are frame-level features. Purple arrows denote outputs which are frame-level probabilities. Red arrows stress update processes of $\theta$. More details in section \ref{MetaSum}.}
	\label{fig:framework}
\end{figure*}

Mathematically, learner is represented by a parametrized function $f_{\theta}$, where ${\theta}$ is a parameter to be learned and is randomly initialized to ${\theta_0}$. To learn (update) ${\theta}$, each iteration is completed by two stages and two training tasks in $Tr\{\tau\}$ are utilized. In the $i$th iteration, ${\theta}$ is updated from ${\theta_{i-1}}$ to ${\theta_i}$ and two training tasks, $\tau_{i}^1$ (the first used training task in the $i$th iteration) and $\tau_{i}^2$ (the second used training task in the $i$th iteration), update ${\theta}$ as follows:

In the first stage, for training task $\tau_{i}^1$, ${\theta}$ can be updated from ${\theta_{i-1}}$ to ${\theta_i^{1} }$ by one gradient descent step and ${\theta_i^{1} }$ can be adjusted to ${\theta_i^{2} }$ on training task $\tau_{i}^1$ by one gradient descent step as well. Theoretically, several adjustments on $\tau_{i}^1$ can be made and ${\theta_i^{n} }$ can be obtained after $n$ adjustments. In the $i$th iteration, one gradient descent update on $\tau_{i}^1$ in the $j$th adjustment takes the form:
\begin{equation}
\left\{ \begin{array}{l}
\theta _i^j = {\theta _{i - 1}} - \alpha {\nabla _{{\theta _{i - 1}}}}{L_{\tau _i^1}}({f_{{\theta _{i - 1}}}}),{\rm{       j = 1}}\\
\theta _i^j = \theta _{_i}^{j - 1} - \alpha {\nabla _{\theta _{i-1}}}{L_{\tau _i^1}}({f_{\theta _{_i}^{j - 1}}}),{\rm{    j = 2,3,}}...{\rm{,n}}
\end{array} \right.,
\label{con:firstLearning}
\end{equation}
where ${\alpha}$ denotes learning rate and is fixed as a hyperparameter. $L_{\tau_{i}^1}(f_{\theta_{i-1}})$ and $L_{\tau_{i}^1}(f_{\theta_{i}^{j-1}})$ are ${L_{1}}$ losses on $\tau_{i}^1$ and states of learner are $\theta_{i-1}$ and $\theta_{i}^{j-1}$ respectively. Specifically, $L_{\tau_{i}^1}(f_{\theta_{i-1}})$ is described as:
\begin{equation}
L_{\tau_{i}^1}(f_{\theta_{i-1}}) = \frac{1}{N} |f_{\theta_{i-1}}(v_{i}^1) - y_{i}^1|,
\end{equation}
where ${N}$ is the number of frames of $v_{i}^1$. $y_{i}^1$ is the annotation of $v_i^1$ which is a score vector with length ${N}$. Here, $f_{\theta_{i-1}}(v_{i}^1)$ denotes output of learner and state of learner is represented by $\theta_{i-1}$. Output $f_{\theta_{i-1}}(v_{i}^1)$ is a score vector which has the same length as the video and the $j$th element in it represents probability of the $j$th frame being selected to summary. Loss $L_{\tau_{i}^1}(f_{\theta_{i}^{j-1}})$ takes the same form of $L_{\tau_{i}^1}(f_{\theta_{i-1}})$ in addition to state of learner changes form $\theta_{i-1}$ to $\theta_{i}^{j-1}$.

The $i$th iteration ends with the second stage where ${\theta_i^{n} }$ is updated on $\tau_{i}^2$ by one gradient descent step:
\begin{equation}
\theta_i = \theta_{i-1} - \beta \nabla_{\theta_{i-1}} L_{\tau_{i}^2} ( f_{\theta_i^{n}} ),
\label{con:secondLearning}
\end{equation}
where $\theta_i^{n}$ denotes state of learner after $n$ adjustments on $\tau_{i}^1$ from $\theta_{i-1}$. $L_{\tau_{i}^2} ( f_{\theta_i^{n}} )$ is ${L_{1}}$ loss on $\tau_{i}^2$ and $\beta$ represents meta learning rate which is fixed as a hyperparameter. $\theta_i$ is updated state of learner after the $i$th iteration.

For simplicity of description, only the $i$th iteration for updating parameter is presented, but multiple iterations in MetaL-TDVS is a straightforward extension as shown in Algorithm \ref{alg::Algo}. By minimizing expected generalization loss of $f_{\theta}$ with respect to ${\theta}$ on  $Val\{\tau\}$, as shown in (\ref{con:endSign}), parameter ${\theta}$ of learner can be obtained. In experiments, we use early stopping strategy and training is stopped when expected generalization loss does not decrease in 800 iterations or maximum iteration (30000) is achieved.
\begin{equation}
	\mathop {\min }\limits_\theta  E\left[ {{L_{Val\{ \tau \} }}\left( {{f_\theta }} \right)} \right].
	\label{con:endSign}
	\end{equation}

Fig. \ref{fig:framework} illustrates process of the $i$th iteration where $n=2$ in detail. As shown, each update of parameter consists of two stages and two training tasks are employed. First, parameter is tuned on the first training task by several gradient descent steps. Then, this update ends with adjustment on the second training task based on tuned parameter. Moreover, MetaL-TDVS is not the special case with batch size of 1. Because each update of MetaL-TDVS contains two stages on two training tasks, and value of $n$ can be any positive integer in theory. Thus, learner is updated by higher order derivative. To simplify the description, all hyperparameters of MetaL-TDVS are represented by $\alpha$, $\beta$ and $n$ where $\alpha$ and $\beta$ denote learning rate and meta learning rate respectively, and $\theta$ is updated on the first used training task $n$ times in the first stage of each iteration.


On the other hand, it can be found from Fig. \ref{fig:framework} and (\ref{con:secondLearning}) that the two-stage learning in each training step forces transcendental task (the first used training task) to provide experiences for next learning (learning on the second used training task and the learning is based on state $\theta_i^n$ which is learned from the first used training task). Associating learning among different tasks in each update is actually propitious to the learning. Because these different tasks essentially have the same nature, that is summarizing video. Moreover, all tasks are not considered in isolation in total learning process, which helps learner reuse previous experiences from different tasks, learn faster, and perform better.

Formulating as a meta learning problem, MetaL-TDVS treats summarizing each video as a single task and forces learner to learn information of task level. The learning which proceeds among tasks (video summarization tasks) makes learner focus more on video summarization task itself rather than only on data. Furthermore, the strategy figured out by learner among tasks is exactly the latent mechanism for video summarization and what the framework intends to excavate. The learning which is from tasks instead of data facilitates exploration of the latent mechanism.

Note that differences between MetaL-TDVS and existing supervised video summarization methods (denoted by ESVSs for simiplicity) can be summarized as follows:

\begin{enumerate}
	\item MetaL-TDVS formulates video summarization as a meta learning problem, but ESVSs mainly formulate it as a subset selection, a structured prediction, or a sequential decision making problem.
	\item In addition to sequential nature of video data, MetaL-TDVS pays more attention to video summarization problem itself; but ESVSs have not statemented this clearly and majority of ESVSs mainly focus on structural or sequential characteristic of video data.
	\item MetaL-TDVS aims to force the specific model (which summarizes video directly) to explicitly explore the mechanism for summarizing video, but ESVSs have not claimed to explore the mechanism unequivocally. 	
\end{enumerate}



\begin{algorithm}
	\caption{MetaL-TDVS to Learn Parameter of Learner}
	\label{alg::Algo}
	\begin{algorithmic}[1]
		\Require \\
		$Tr\{\tau\},Val\{\tau\},Te\{\tau\}$: training, validation, test task set;  \\
		$\alpha$: learning rate;          \\
		$\beta$: meta learning rate;          \\
		$n$: number of times adjusted on the first used training task;
		\Ensure \\
		$\theta$ : parameter of learner; \\
		\State Initialize $\theta=\theta_0$ and $i=1$;
		\While {$not\ done$}
		\State Sample two training tasks $\tau_{i}^1$ and $\tau_{i}^2$ from $Tr\{\tau\}$;
		\For {$j=1$ to $n$}
		\If {$j == 1$}
		\State  $\theta_{i}^{j} \gets \theta_{i-1} - \alpha \nabla_{\theta_{i-1}} L_{\tau_{i}^1}(f_{\theta_{i-1}})$;
		\EndIf
		\If {$j > 1$}
		\State $\theta_{i}^{j} \gets \theta_{i}^{j-1} - \alpha \nabla_{\theta_{i-1}} L_{\tau_{i}^1}(f_{\theta_{i}^{j-1}})$;
		\EndIf
		\EndFor
		\State $\theta_i \gets \theta_{i-1} - \beta \nabla_{\theta_{i-1}} L_{\tau_{i}^2} ( f_{\theta_i^{n}} )$;
		\State $i \gets i+1$;
		\EndWhile
	\end{algorithmic}
\end{algorithm}

\subsection{Specific Model for the Learner}
\label{modelForLearner}

To show effectiveness of MetaL-TDVS, we consider two types of features. The first one is deep feature extracted from output of the penultimate layer of GoogLeNet \cite{szegedy2015going}. By using this feature extraction method, each frame of input video is encoded into a 1024-dimensional feature descriptor.  The second one is traditional feature consisting of four image descriptors: color histograms, GIST, HOG and dense SIFT. Color histograms are computed from RGB images and all the other features are extracted on gray scale images.

On the other hand, because video summarizaion is made based on storyline which progresses through entire video, the sequential or structural nature of video data is also of great importance to effectively address video summarization problem. Thus, ways that only rely on visual cues and do not take consideration of temporal relation across frames are not such qualified to summarize video. To get an ideal video summary, highlevel semantic understanding about video over a long-range temporal span needs to be taken into account. So we employ vsLSTM \cite{zhang2016video} to implement learner in MetaL-TDVS.

\begin{figure}[t]
	\centering
	\includegraphics[width=9cm]{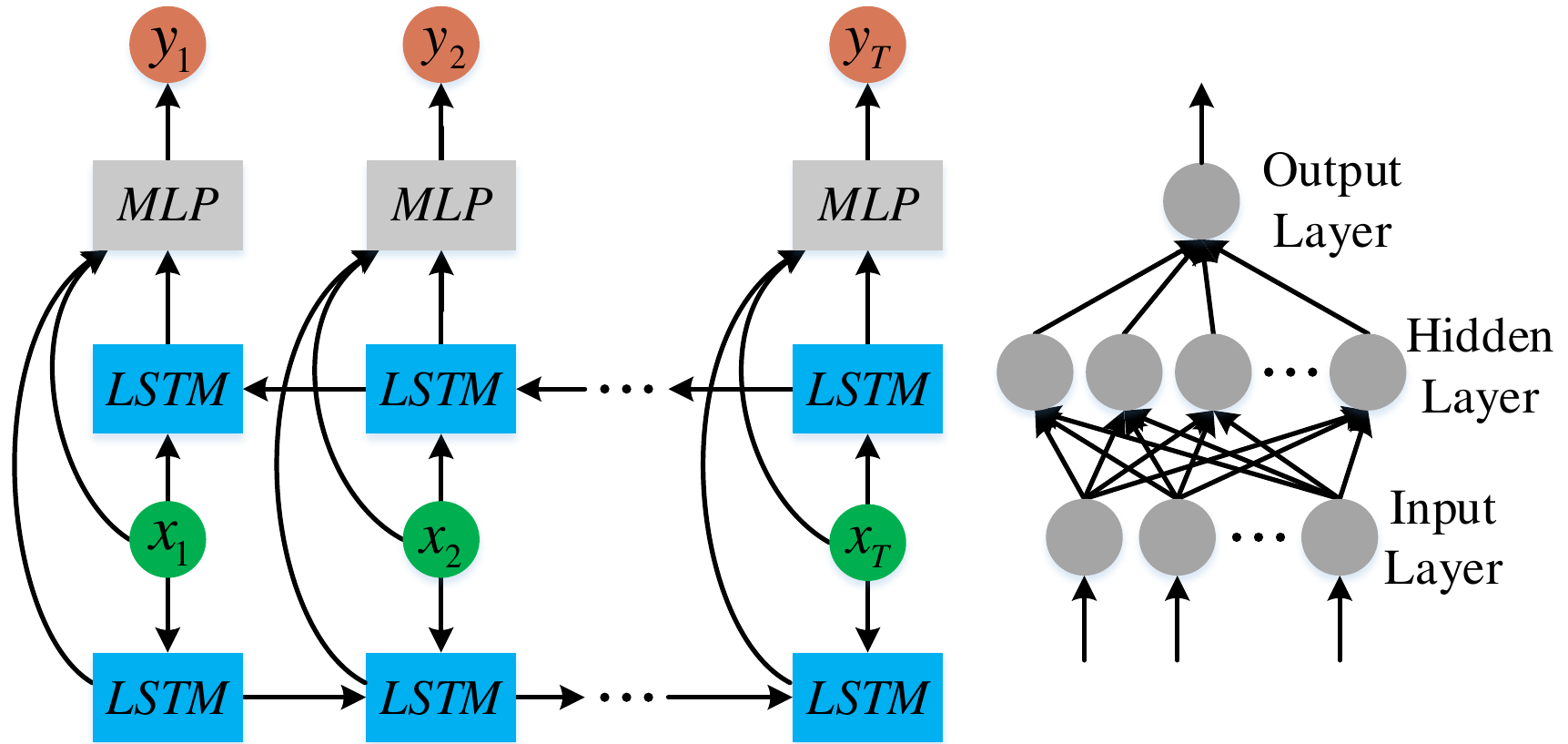}
	\setlength{\abovecaptionskip}{-0.5cm}
	\setlength{\belowcaptionskip}{-0.8cm}
	\caption{Structures of the employed learner (vsLSTM) and MLP. Left side shows network structure of vsLSTM where green circles represent input features of video frames and brown ones are outputs (probabilities to be selected). Blue rectangles indicate LSTM units and gray ones are MLPs. Right side shows structure of one MLP which has an input layer, one hidden layer and an output layer.}
	\label{fig:vslstm}
\end{figure}
The vsLSTM consists of bidirectional LSTM layers \cite{graves2005framewise} and one multi-layer perceptron (MLP) layer. For clarity, we plot the structure of the employed learner (vsLSTM), as well as the structure of  MLP, in Fig. \ref{fig:vslstm}. There is no direct interaction between forward and backward LSTM layers. Combining hidden states of these two LSTM layers and features of video frames, MLPs are all implemented by one-hidden-layer and are utilized to compute probabilities of frames. Hidden units and output layers of MLPs are all activated by sigmoid activation function. The size of hidden layers of MLPs, the number of hidden units of each unidirectional LSTM as well as the output dimension of MLPs are all 256.

\section{Experiments}	
\label{experiment}

This section firstly presents detailed descriptions of experimental setups, then various experiments are carried out to demonstrate the efficiency and superiority of MetaL-TDVS.

\subsection{Experimental Setups}

\subsubsection{Datasets} Performance of MetaL-TDVS is evaluated on SumMe \cite{gygli2014creating} and TVSum \cite{song2015tvsum}.

There are 25 user videos in SumMe and events recorded by these videos are multifarious, such as sports and holidays. Both ego-centric and third-person camera are included, and contents expressed are diverse. Video lengths range from 1.5 to 6.5 minutes and provided labels are frame-level importance scores. TVSum consists of 50 videos downloaded from YouTube and videos are organized into groups with a key-word as topic of each group. Selected from 10 categories in TRECVid Multimedia Event Detection (MED), the 50 videos are organized into 10 topics (5 videos per topic) and lengths of them are in range of 1 to 5 minutes. Videos in TVSum include first-person and third-person camera and contents are extremely diverse. Labels are frame-level importance scores.

To invistigate generalization ability of learned model and combat the need of huge amount of annotated data, the other two datasets, Youtube \cite{de2011vsumm} and Open Video Project (OVP) \cite{OVP}, are also utilized. Youtube includes 50 videos collected from websites and contents include news and sports. Video lengths vary from 1 to 10 minutes and annotations provided are multiple user-annotated subsets of keyframes for each video. For OVP, we utilize the same 50 videos as \cite{de2011vsumm}. Videos are from various genres such as documentary and educational, and their lengths are form 1 to 4 minutes.

\subsubsection{Evaluation Metrics}

To make a fair comparison, we use keyshot-based metrics proposed in \cite{zhang2016video} for evaluation which follow protocols in \cite{gygli2014creating,gygli2015video} as well.

Suppose $A$ is the generated keyshot-based summary and $B$ is the human-annotated keyshots. Precision ($P$) and recall ($R$) against human-annotated summary $B$ are computed according to temporal overlap between them:
\begin{equation}
\setlength{\abovedisplayskip}{2pt}
\setlength{\belowdisplayskip}{3pt}
P = \frac{{A \cap B}}{A}, R = \frac{{A \cap B}}{B},
\end{equation}
the finally used harmonic mean F-score ($F$) is computed as:
\begin{equation}
\setlength{\abovedisplayskip}{2pt}
\setlength{\belowdisplayskip}{3pt}
F = 2P \times R / (P+R) \times 100 \%.
\end{equation}

\subsubsection{Implementation Details}

To generate key frames or key shots, we follow methods described in \cite{zhang2016video}. Videos are temporally segmented into disjoint intervals by kernel temporal segmenation (KTS) according to frame scores. Based on importance score of each interval (average importance score of frames in the interval), resulting intervals are ranked. Summary consists of keyshots selected from ranked intervals and total duration of summary is less than 15\% of input video.
To obtain a single ground-truth set when there are multiple human annotations, we use the algorithm proposed in \cite{gong2014diverse}. For each video with multiple annotations, single ground-truth set $y^*$ is initialized to be empty and one frame $f$ is added to $y^*$ by maximizing (\ref{con:multiple}):
\begin{equation}
{y^{\rm{*}}} \leftarrow {y^{\rm{*}}} \cup \mathop{\arg\max}_{f} \mathop \sum \limits_{j = 1}^m {F_{{y^{\rm{*}}} \cup f,{y_{j}}}},
\label{con:multiple}
\end{equation}
where $m$ is the number of annotations and ${y_{j}}$ denotes the $j$th annotation. $F_{{y^{\rm{*}}} \cup f,{y_{j}}}$ represents F-score of $y^{*} \cup f$ and $y_{j}$. Frames not in $y^*$ can be iteratively added to $y^*$ until there is no frame increases the F-score.

The way to split datasets into training, validation, testing sets is referenced from \cite{zhang2016video}. We follow the ``Transfer'' way in learning: for a given dataset (SumMe or TVSum), the other three datasets are utilized for training and validation, then the learned model is tested on that dataset. This way allows us to verify generalization ability of learned model on an unseen dataset. We run it for each testing fold 5 times and average results are computed as final results.

\subsection{Results}

In this subsection, we investigate the sensitivity of hyperparameters and structures, followed by comparisons with representative methods.

\subsubsection{Sensitivity Evaluation of Hyperparameters}
\label{Exp:hyperparameters}
\begin{figure}
	\centering
	\includegraphics[width=.50\textwidth]{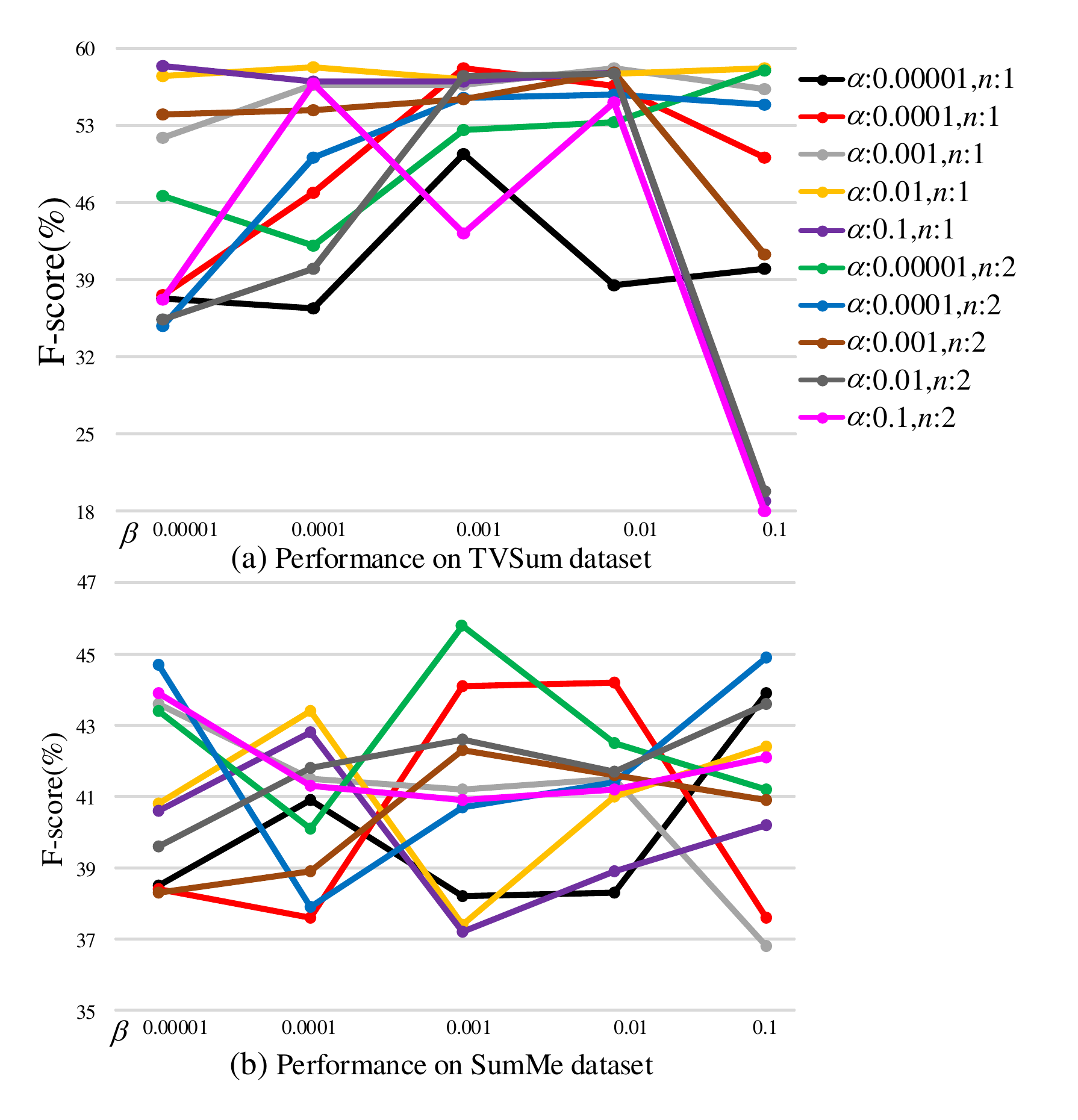}
	\setlength{\abovecaptionskip}{-0.8cm}
	\setlength{\belowcaptionskip}{-0.8cm}
	\caption{Results of MetaL-TDVS with different hyperparameters.}
	\label{fig:diffeParam}
\end{figure}

Performances of MetaL-TDVS with different hyperparameters ($\alpha$, $\beta$, $n$) are evaluated and shown in Fig. \ref{fig:diffeParam} where $\alpha$ and $\beta$ can be 0.1, 0.01, 0.001, 0.0001, and 0.00001. Due to limitation of video memory, we test performances when $n$ is equal to 1 or 2.

\begin{figure}[t]
	\centering
	\includegraphics[width=.50\textwidth]{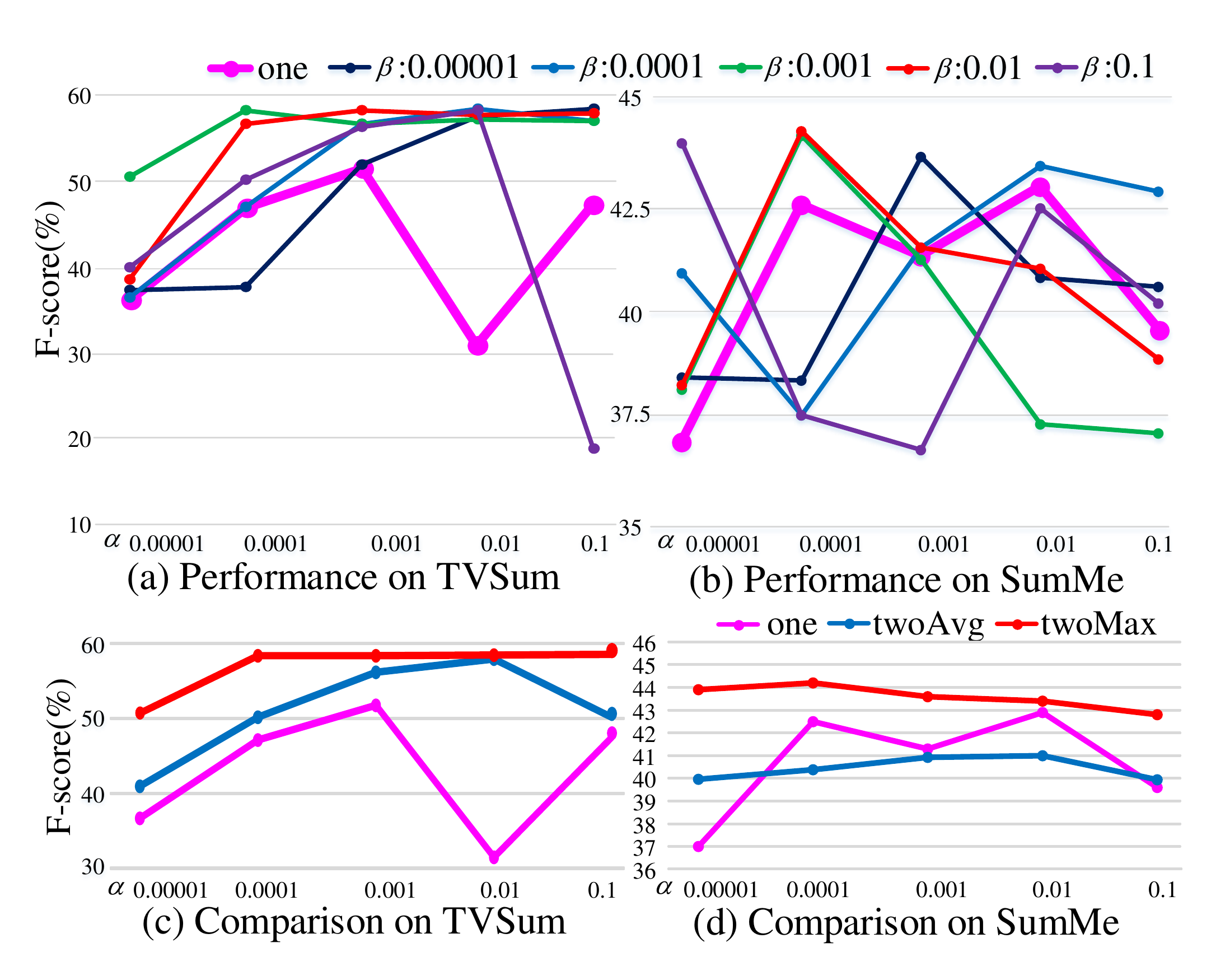}
	\setlength{\abovecaptionskip}{-0.8cm}
	\setlength{\belowcaptionskip}{-0.8cm}
	\caption{Performance and comparison of with and without the second stage. ``one'' denotes without the second stage (only the first stage with hyperparameters $\alpha$ and $n$). All of others in (a) and (b) represent MetaL-TDVS with different $\beta$ and have two stages with hyperparameters $\alpha$, $\beta$ and $n$. Both ``twoAvg'' and ``twoMax'' are statistics for models with the second stage. In specific, ``twoAvg'' and ``twoMax'' represent the average and maximum with respect to $\beta$ when $\alpha$ and $n$ specified.}
	\label{fig:oneStage}
\end{figure}

\begin{figure}[t]
	\centering
	\includegraphics[width=.50\textwidth]{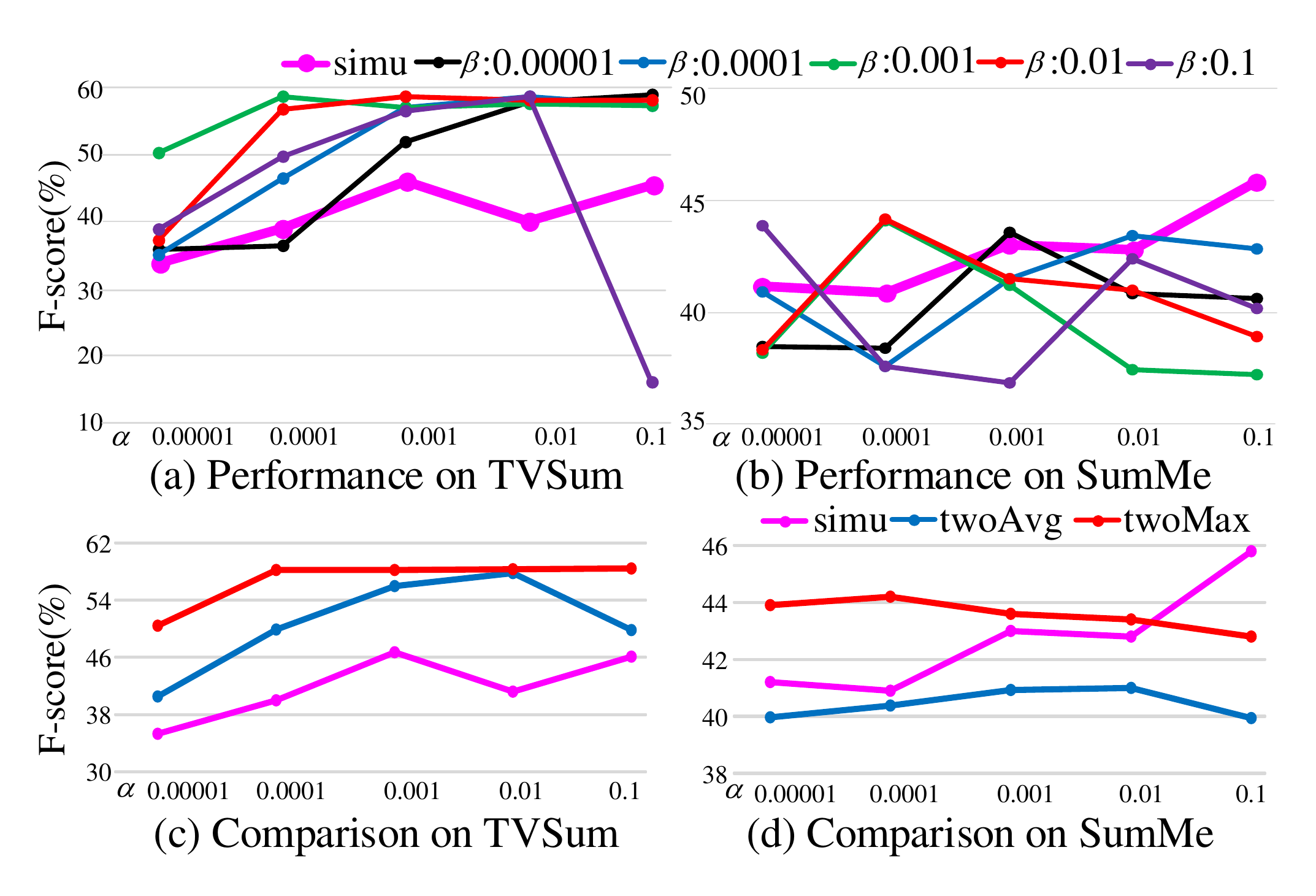}
	\setlength{\abovecaptionskip}{-0.8cm}
	\setlength{\belowcaptionskip}{-0.8cm}
	\caption{Performance and comparison of simultaneously learning and successively learning. ``simu'' denotes training on two tasks simultaneously in each iteration. All of others represent MetaL-TDVS with different $\beta$ and train on two tasks successively in each iteration. Both ``twoAvg'' and ``twoMax'' are statistics for models where training on two tasks successively in each iteration as shown in Fig. \ref{fig:framework}. ``twoAvg'' and ``twoMax'' are the same meaning as in Fig. \ref{fig:oneStage}.}
	\label{fig:simu}
\end{figure}

\begin{figure*}[t]
	\centering
	\includegraphics[width=18cm]{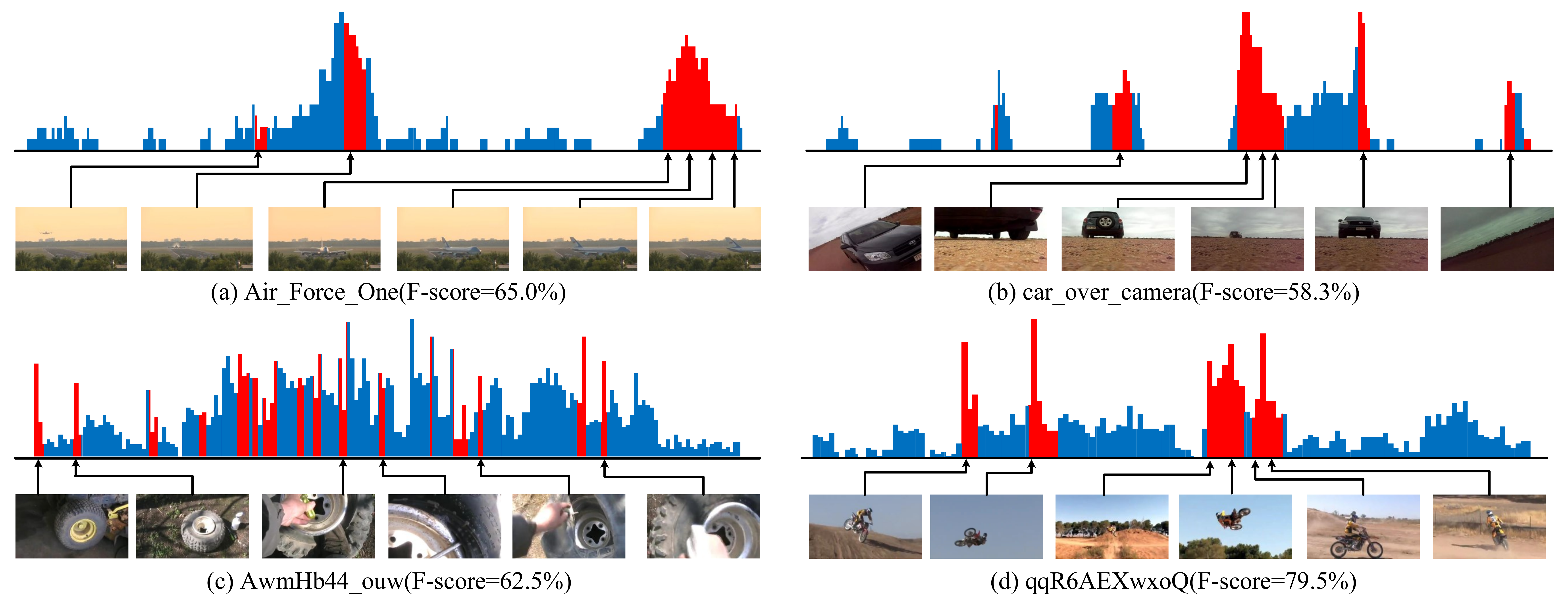}
	\setlength{\abovecaptionskip}{-0.8cm}
	\setlength{\belowcaptionskip}{-0.8cm}
	\caption{ Exemplar summaries (red intervals) from four sample videos with ground-truth importance scores (blue background).}
	\label{fig:visualization}
\end{figure*}

It can be seen that for each $\beta$, MetaL-TDVS with different $\alpha$s and different $n$s gives different results, and different performances are presented on the two datasets. For instance, performance of MetaL-TDVS represented by pink polyline (with $\alpha$=0.1, $n$=2) on TVSum shocks drastically with the change of $\beta$, but its counterpart (pink polyline on SumMe) has completely distinct trend. This change trend discrepancy on these two datasets also occurs when $\alpha$=0.1, $n$=1 (purple polyline), $\alpha$=0.00001, $n$=1 (black polyline), $\alpha$=0.01, $n$=1 (orange polyline) etc. Based on results on different hyperparameters, the one with $\beta$=0.001, $\alpha$=0.0001, $n$=1 (the third point of red polyline) has better generalization ability (performs well on both the two datasets). Because though the one with $\beta$=0.001, $\alpha$=0.00001, $n$=2 (the third point of green polyline) performs best on SumMe, it gets poor results on TVSum (the first point of blue polyline with $\beta$=0.00001, $\alpha$=0.0001, $n$=2 the same). Though the one with $\beta$=0.00001, $\alpha$=0.1, $n$=1 (the first point of purple polyline) gets promising results on TVSum, it performs poor on SumMe (the second and the last points of orange polyline with $\beta$=0.0001, $\alpha$=0.01, $n$=1 and $\beta$=0.1, $\alpha$=0.01, $n$=1 the same). But MetaL-TDVS has similar performance change trend on both these two datasets with $\alpha$=0.0001 and n=1 (red polyline). Thus, to get better generalization ability, $\alpha$ is set to 0.0001 and $n$ set to 1 in reported results. Based on the two fixed hyperparameters, $\beta$ is 0.001 since better results are got on both the two datasets.

\setlength{\tabcolsep}{4pt}
\begin{table}
	\begin{center}
		\caption{
			Performance comparison (F-score\ \%) with seven state-of-the-art methods. Best results are renoted in bold.
		}
		\setlength{\belowcaptionskip}{-0.5cm}
		\label{table:comWithSuper}
		\begin{tabular}{ccc}
			\hline
			Method                            &      TVSum        &      SumMe   \\
			\hline
			Gygli \textit{et al.} \cite{gygli2015video}             &       -       &       39.7          \\
			\hline
			vsLSTM \ \cite{zhang2016video}            &    56.9       &     40.7     \\
			\hline
			Zhang \textit{et al.} \cite{zhang2016summary}	      &   -      &        40.9     \\
			\hline
			SUM-GA${\rm {N_{sup}}}$ \cite{mahasseni2017unsupervised} &  56.3  &   41.7   \\
			\hline
			DSSE \cite{yuan2017video} & 57.0   &  -    \\
			\hline
			DR-DS${\rm {N_{sup}}}$ \cite{zhou2017deep}              &    58.1    & 42.1 \\
			\hline
			Li \textit{et al.} \cite{li2017general}             & 52.7  &    43.1  \\
			\hline
			MetaL-TDVS (ours)    &   \textbf{58.2}   &  \textbf{44.1} \\
			\hline
		\end{tabular}
	\end{center}
\end{table}
\setlength{\tabcolsep}{1.4pt}

\subsubsection{Performance Evaluation on Different Structures}

Because in \ref{Exp:hyperparameters}, $\alpha$=$0.0001$, $\beta$=$0.001$, $n$=$1$ are selected as the final hyperparameters, experiments for different structures are done with $n$=1 (only $\alpha$ and $\beta$ are variable hyperparameters).

To confirm whether the second stage improves performance, frameworks with and without the second stage are tested on SumMe and TVSum. Results are shown in Fig. \ref{fig:oneStage}.(a) and (b) where pink polylines denote without the second stage. All of others have the second stage and different colors represent frameworks with different $\beta$.

It can be seen that for each $\alpha$, there is at least one $\beta$ making the framework with two stages outperform ``one'' and mostly a large margin. The superiority of two stages is extremely obvious on TVSum. Moreover, Fig. \ref{fig:oneStage}.(c) and (d) compares ``one'' with the average and maximum F-scores of frameworks with two stages. ``twoAvg'' represents with the second stage and the average is computed as (when $\alpha = \alpha_i$):
\begin{equation}
\setlength{\abovedisplayskip}{2pt}
\setlength{\belowdisplayskip}{3pt}
{\rm{twoAv}}{{\rm{g}}_{\alpha_i}} = \frac{1}{5}\sum\limits_{\beta_j} {F_{{\alpha_i},{\beta_j}}},
\label{equ:avg}
\end{equation}
where ${F_{{\alpha_i},{\beta_j}}}$ is the F-score of MetaL-TDVS with $\alpha$=$\alpha_i$, $\beta$=$\beta_j$. $\beta_j$ are elements in set $\{0.1,0.01,0.001,0.0001,0.00001\}$, 5 is the number of elements in the set. ${\rm{twoAv}}{{\rm{g}}_{\alpha_i}}$ is the point with $\alpha$=$\alpha_i$ on the corresponding blue polyline. ``twoMax'' indicates with the second stage and the maximum is computed as (when $\alpha = \alpha_i$):
\begin{equation}
\setlength{\abovedisplayskip}{2pt}
\setlength{\belowdisplayskip}{3pt}
{\rm{twoMa}}{{\rm{x}}_{\alpha_i}} = \mathop {\max }\limits_{\beta_j} {F_{{\alpha_i},{\beta_j}}},
\label{equ:max}
\end{equation}
where ${F_{{\alpha_i},{\beta_j}}}$ and $\beta_j$ are the same as in (\ref{equ:avg}), ${\rm{twoMa}}{{\rm{x}}_{\alpha_i}}$ denotes the point with $\alpha$=$\alpha_i$ on the corresponding red polyline.
	
Obviously, ``twoMax'' are better than ``one'' on both the two datasets. Though only two of five points on ``twoAvg'' better than ``one'' (one slightly bad and two visibly poorer than ``one'') on SumMe, all points of ``twoAvg'' better than ``one'' on TVSum. There are still few points on ``twoAvg'' worse than ``one'' because the average is computed on many values of $\beta$ (0.1,0.01,0.001,0.0001,0.00001) when $\alpha$ specified and there may exist some cases where performances are bad enough due to inappropriate $\beta$. Overall, we can state that the second stage really improves performance and mostly a large margin.

To validate whether successively learning on two tasks performs better, frameworks of simultaneously and successively training on two tasks (in each iteration) are tested. Results are shown in Fig. \ref{fig:simu}.(a) and (b) where pink polylines denote training on two tasks simultaneously. All of others indicate training successively (as shown in Fig. \ref{fig:framework}) with different $\beta$.

Evidently, there exists at least one $\beta$ for each $\alpha$ making successively learning outperforms ``simu'' except $\alpha$=0.1 on SumMe, and margins are visibly large in most cases. Furthermore, Fig. \ref{fig:simu}.(c) and (d) (where ``twoAvg'' and ``twoMax'' have similar meaning as in Fig. \ref{fig:oneStage}.(c) and (d) except ${F_{{\alpha_i},{\beta_j}}}$ from Fig. \ref{fig:simu}.(a) and (b) rather than \ref{fig:oneStage}.(a) and (b)) makes comparison between ``simu'' with the average and maximum F-scores of frameworks where training on two tasks successively. Though ``twoAvg'' performs poorer than ``simu'' on SumMe, almost all points on ``twoMax'' outperform ``simu'' distinctly, and ``twoAvg'' on TVSum outperforms ``simu'' a large margin. Thus, it is reasonable to say that training on two tasks successively performs better than simultaneously.

\subsubsection{Comparisons With Representative Methods}

Table \ref{table:comWithSuper} summarizes performance of MetaL-TDVS and makes comparison with seven state-of-the-art supervised methods. For competitors, published results are directly used. Furthermore, we only compare with supervised methods since supervised approaches perform better than unsupervised ones to a certain extent with the help of annotations. Specifically, there are several models proposed in \cite{mahasseni2017unsupervised}, but we only make comparison with its supervised one which performs best in all its proposed models (both unsupervised and supervised). And \cite{zhou2017deep} the same.

Shown in Table \ref{table:comWithSuper}, MetaL-TDVS performs better than competitors on SumMe and TVSum. Despite MetaL-TDVS performs slightly better than DR-DSN$\rm_{sup}$ on TVSum, there are 2.0 percentage increases in performance on SumMe. In addition, MetaL-TDVS outperforms the approach proposed by Li \emph{et al}. and there are 1.0 and 5.5 percentage increases on SumMe and TVSum respectively. On the two datasets, there are 1.4 and 3.3 percentage points better than the vsLSTM (the same ``Transfer'' learning settings as MetaL-TDVS) respectively which is the model implementing learner in MetaL-TDVS.

As is expected, experimental results demonstrate the superiority and effectiveness of MetaL-TDVS, and also indicate the way of meta learning is suitable to summarize video.

To promote generalization ability and make ideal video summaries, model is supposed to learn how to summarize video. Thus, what the model learned form processes of summarizing other videos, that is how to summarize video, is essentially what it needs to summarize new ones. In fact, this is in complete accord with the idea of meta learning which is making better use of the experience and knowledge learned from other tasks (summarize other videos) to handle new ones (summarize new videos). Therefore, meta learning is a reasonable way to summarize video and this is verified by experimental results. Besides, laying more stress on video summarization problem itself rather than only on sequential or structural video data, MetaL-TDVS forces the model to explicitly explore the mechanism for summarizing video among tasks and is superior in terms of generalization ability.

\setlength{\tabcolsep}{4pt}
\begin{table}
	\begin{center}
		\caption{
			Performance comparison (F-score\ \%) with state-of-the-art supervised methods when using traditional features. Best results are renoted in bold.
		}
		\setlength{\belowcaptionskip}{-0.5cm}
		\label{table:perforOnShall}
		\begin{tabular}{ccc}
			\hline
			Method                         &   TVSum   &   SumMe   \\
			\hline
			SUM-GA${\rm {N_{sup}}}$ \cite{mahasseni2017unsupervised}&\textbf{59.5}&39.5\\
			\hline
			dppLSTM\ \cite{zhang2016video} & 57.9 & 40.7 \\
			\hline
			MetaL-TDVS (ours)   &  57.9  &   \textbf{43.5} \\
			\hline
		\end{tabular}
	\end{center}
\end{table}
\setlength{\tabcolsep}{1.4pt}

\setlength{\tabcolsep}{4pt}
\begin{table}
	\begin{center}
		\caption{
			Time comparison (frames per second, fps) with two models. Best results are renoted in bold.
		}
		\setlength{\belowcaptionskip}{-0.5cm}
		\label{table:t}
		\begin{tabular}{cccc}
			\hline
			Method  & infer  & final & total \\
			\hline
			vsLSTM\ \cite{zhang2016video} & 2040 & 3470 & 1285\\
			\hline
			dppLSTM\ \cite{zhang2016video} & 1956 & 1306 & 783 \\
			\hline
			MetaL-TDVS (ours)  & \textbf{2050} & \textbf{49751} & \textbf{1969} \\
			\hline
		\end{tabular}
	\end{center}
\end{table}
\setlength{\tabcolsep}{1.4pt}

\subsubsection{Generalization on Non-Deep Features}

Generalization ability of MetaL-TDVS to non-deep features is demonstrated by evaluating its performance with shallow features as utilized in \cite{song2015tvsum}. Table \ref{table:perforOnShall} summarizes performances of MetaL-TDVS and some state-of-the-art supervised methods where only shallow features are used.

It can be seen that MetaL-TDVS performs comparable to competitors. On TVSum, SUM-GA${\rm {N_{sup}}}$ performs best and it is 1.6 percentage better than MetaL-TDVS. But on SumMe, MetaL-TDVS outperforms the competitors, and there are 4.0 and 2.8 percentage increase than SUM-GA${\rm {N_{sup}}}$ and dppLSTM (``Transfer'' learning settings) respectively. Promising results demonstrate the robustness on non-deep features of MetaL-TDVS.

\subsubsection{Qualitative Results}
To show performance of MetaL-TDVS intuitively, selected frames on four videos (Air\_Force\_One and car\_over\_camera of SumMe, AwmHb44\_ouw and qqR6AEXwxoQ of TVSum) are demonstrated in Fig. \ref{fig:visualization}. Blue blocks represent ground-truth frame-level importance scores and the ones selected by MetaL-TDVS are marked red. Colored regions are several selected frames in video summaries. Despite some variations, it can be seen that MetaL-TDVS is able to extract frames with high importance and discard the ones which do not contain enough valuable information.

\subsubsection{Performance on Specific Types of Videos}
The proposed MetaL-TDVS is a generic video summarization method and not specific to certain types of video. To see how it will behave on some specific types of videos, we test it on fast moving football matches, slow video, and long video (such as full 3hrs).

For fast moving football matches, we selected 17 (v71 to v87) videos in Youtube which are all about football matches and the lengths range from 1min to 10min. Trained on OVP, SumMe and TVSum, learner is tested on these videos and average precision, recall and F-score are 48.53\%, 42.16\%, and 41.1\% respectively. For slow video, we selected four videos in SumMe (Air\_Force\_One, Bus\_in\_Rock\_Tunnel, Cockpit\_Landing, and St Maarten Landing) where Air\_Force\_One records process of landing a plane from a fixed perspective; Bus\_in\_Rock\_Tunnel shows how a bus passing through a tunnel (moves very very slowly); Cockpit\_Landing records view of birds eye of the ground in airplane and then the airplane lands; St Maarten Landing shows process of plane landing near the beach. Trained on OVP, Youtube, and TVSum, learner is tested on these videos and average precision, recall and F-score are 55.15\%, 54.08\%, and 57.28\% respectively. It can be seen that MetaL-TDVS performs slightly better on slow videos than fast moving ones.
	
For long videos, we use four videos (P01-P04) in \cite{lee2012discovering}. Durations of these videos (P01-P04) are 3 hours 51 minutes 51 seconds, 5 hours 7 minutes 37 seconds, 2 hours 59 minutes 16 seconds, and 4 hours 59 minutes respectively. All of these videos record equipment wearers' daily life where P01-P03 mainly include shopping, eating, driving, cocking and interacting with others while P04 shows working indoors and outdoors with computer. Because the method proposed in \cite{lee2012discovering} is important people and objects based (summary is made according to the detection response in each frame and each video segment), the ground truth provided are pixel-wise which can not be used by MetaL-TDVS due to basic modeling of video summarization is not the same. Thus, we shorten these videos by model trained on SumMe and TVSum, and provide summarized result videos (which are available by https://pan.baidu.com/s/1DsSd0h8OcU3nil8RY3UXSQ\#list/pa th=\%2F) rather than quantitative measures such as F-score, precision and recall. In experiments, videos (P01-P04) are all sampled to 1500 frames around due to classical sample way (select 2 frames per second) produces too many sampled frames which can not be totally loaded into our video memory. Frames in summary generate result summarized video at a frame rate of 2fps. From result videos, it can be seen that main contents of each long video are captured which can demonstrate the effectiveness of MetaL-TDVS to a certain extent.

\subsubsection{Time Comparison}
Table \ref{table:t} shows time comparison between MetaL-TDVS and vsLSTM, dppLSTM where ``infer'' stage with frame features as input and importance scores as output, and ``final'' stage with importance scores as input and binary labels of being selected or non-selected as outputs, ``total'' with frame features as input and binary labels of frames as output. It is evident MetaL-TDVS is faster than competitors.
\section{Conclusions}

\label{conclusion}

In this paper, we reformulate summarizing video as a meta learning problem and propose a novel and effective method MetaL-TDVS. MetaL-TDVS views summarizing each video as a single task and the learning proceeds among tasks. This way of learning makes learner focus more on video summarization problem itself and facilitates exploring for the latent mechanism of summarizing video. Experimental results reveal that MetaL-TDVS is effective and outperforms recently state-of-the-art methods including GAN based and deep reinforcement learning based methods. So meta learning is suitable to summarize video. In future, we will further explore to summarize video with meta learning. On the one hand, we intend to design more suitable network models for learner to better capture intrinsic sequential and structural characteristic of video data (from perspective of video data); on the other hand, we plan to devise more superior meta learners (training frameworks or real models) to force learner to better explore the mechanism for summarizing video at the same time (from perspective of video summarization task level). Besides, some foreground extraction \cite{li2018superpixel} or manifold learning \cite{li2018patch} methods may be used for improving the performance of video summarization as well.
	



{\bibliographystyle{IEEEtran}
\bibliography{reference}
} 

{	
{Xuelong Li} (M'02-SM'07-F'12) is a full professor with the School of Computer Science and the Center for OPTical IMagery Analysis and Learning (OPTIMAL), Northwestern Polytechnical University, Xi'an 710072, P. R. China.

\begin{IEEEbiography}[{\includegraphics[width=1in,height=1.25in,clip,keepaspectratio]{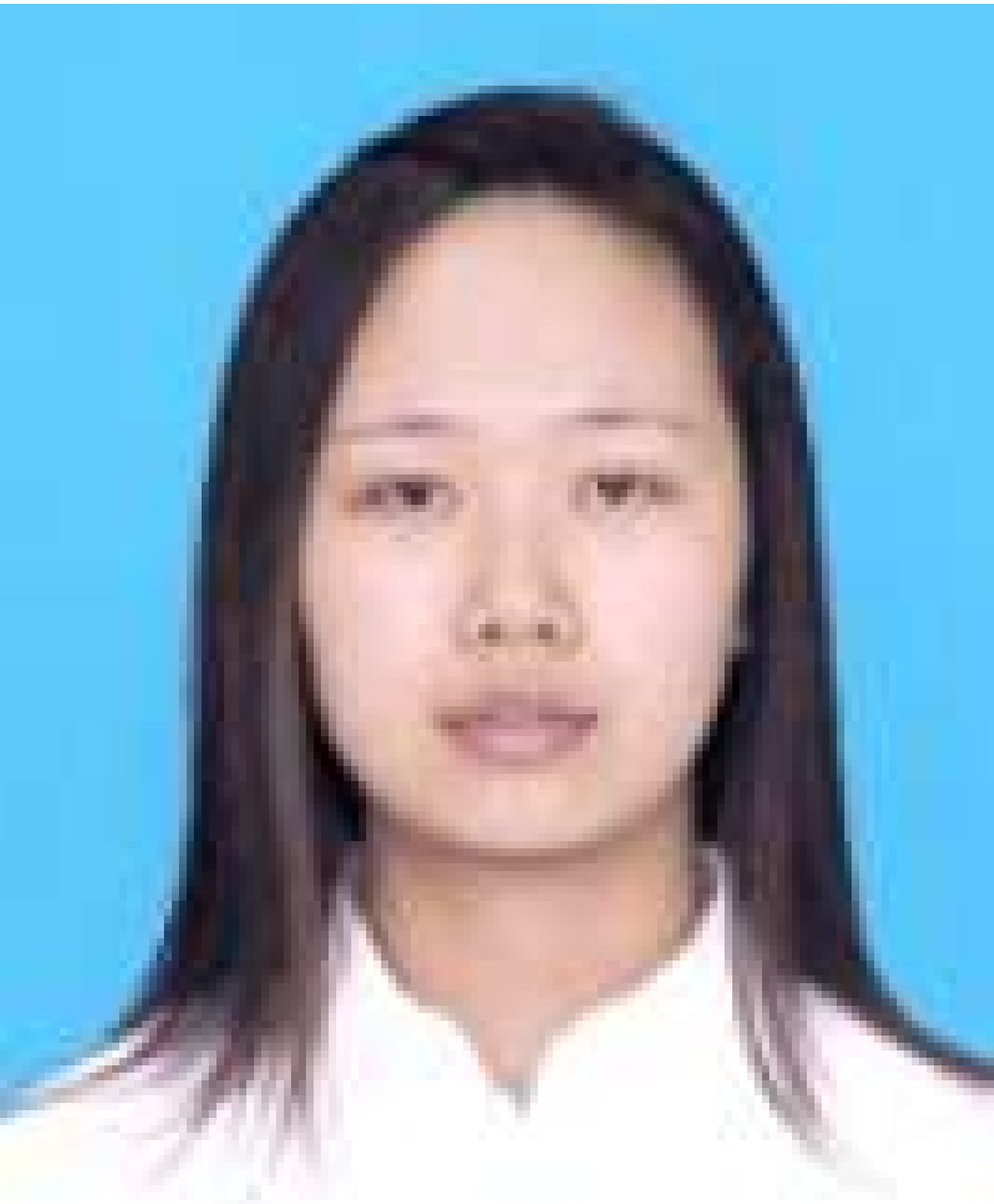}}]
{Hongli Li} is working toward the Ph. D. degree in the School of Computer Science and the Center for OPTical IMagery Analysis and Learning (OPTIMAL), Northwestern Polytechnical University, Xi'an 710072, P. R. China.
\end{IEEEbiography}

\begin{IEEEbiography}[{\includegraphics[width=1in,height=1.25in,clip,keepaspectratio]{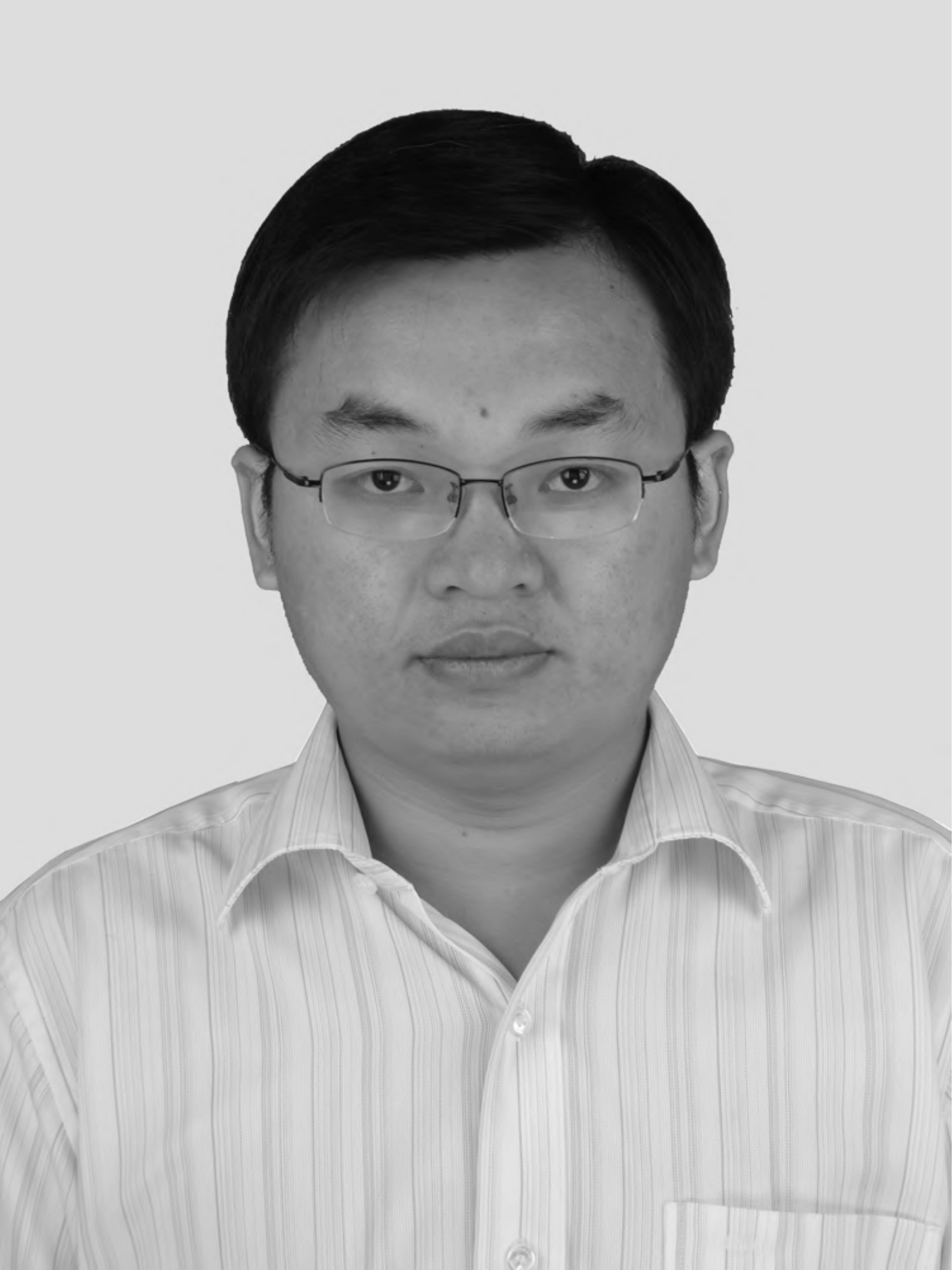}}]
{Yongsheng Dong} (SM'19) received his Ph. D. degree in applied mathematics from Peking University in 2012. He was a postdoctoral research fellow with the Center for Optical Imagery Analysis and Learning, Xi'an Institute of Optics and Precision Mechanics, Chinese Academy of Sciences, Xi'an, P. R. China from 2013 to 2016. From 2016 to 2017, he was a visiting research fellow at the School of Computer Science and Engineering, Nanyang Technological University, Singapore. He is currently an associate professor with the School of Information Engineering, Henan University of Science and Technology, P. R. China. His current research interests include pattern recognition, machine learning, and computer vision.

He has authored and co-authored over 30 papers at famous journals and conferences, including IEEE TIP, IEEE TNNLS, IEEE TCYB, IEEE TCSVT, IEEE SPL and ACM TIST. He is an associate editor of Neurocomputing.
\end{IEEEbiography}}

\end{document}